\documentclass{article}

\usepackage{arxiv}

\usepackage[utf8]{inputenc} 
\usepackage[T1]{fontenc}    
\usepackage{hyperref}       
\usepackage{url}            
\usepackage{booktabs}       
\usepackage{nicefrac}       
\usepackage{microtype}      
\usepackage{lipsum}		
\usepackage{natbib}
\usepackage{doi}
\setcitestyle{numbers,square}
\usepackage{amsmath,amssymb,amsfonts}
\usepackage{graphicx,color}
\usepackage{textcomp}
\usepackage{xcolor}
\usepackage{algorithm,algorithmic}
\usepackage{threeparttable} 
\usepackage{multirow}

\title{ViBED-Net: Video Based Engagement Detection Network Using Face-Aware and Scene-Aware Spatiotemporal Cues\thanks{{This paper is currently under review at a publisher.}}}

\author{ \href{https://orcid.org/0009-0007-9888-6880}{\includegraphics[scale=0.06]{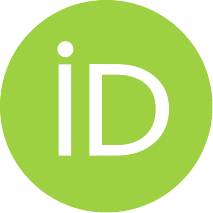}\hspace{1mm}Prateek Gothwal} \\
	Computer Science \& Engineering,\\
	University of Colorado Denver,\\
	Colorado, CO 80204, \\
	\texttt{prateek.gothwal@ucdenver.edu} \\
	\And
\href{https://orcid.org/0000-0002-6967-3807}{\includegraphics[scale=0.06]{orcid.pdf}	Deeptimaan Banaerjee} \\
	Computer Science \& Engineering,\\
	University of Colorado Denver,\\
	Colorado, CO 80204, \\
	\texttt{deeptimaan.banerjee@ucdenver.edu}
	\AND
	\href{https://orcid.org/0000-0002-7446-4639}{\includegraphics[scale=0.06]{orcid.pdf}\hspace{1mm}Ashis Kumer Biswas} \\
	Computer Science \& Engineering,\\
	University of Colorado Denver,\\
	Colorado, CO 80204, \\
	\texttt{ashis.biswas@ucdenver.edu}
}

\date{October 20, 2025}

\renewcommand{\shorttitle}{ViBED-Net: Video Based Engagement Detection Network}

\hypersetup{
pdftitle={ViBED-Net: Video Based Engagement Detection Network},
pdfsubject={AI},
pdfauthor={Prateek Gothwal, Deeptimaan Banerjee, Ashis Kumer Biswas},
pdfkeywords={Student Engagement Detection, EfficientNet, Spatiotemporal Modeling, Dual-Stream Architecture}
}

\begin{document}
\maketitle

\begin{abstract}
Engagement detection in online learning environments is vital for improving student outcomes and personalizing instruction. We present ViBED-Net (Video-Based Engagement Detection Network), a novel deep learning framework designed to assess student engagement from video data using a dual-stream architecture. ViBED-Net captures both facial expressions and full-scene context by processing facial crops and entire video frames through EfficientNetV2 for spatial feature extraction. These features are then analyzed over time using two temporal modeling strategies: Long Short-Term Memory (LSTM) networks and Transformer encoders. Our model is evaluated on the DAiSEE dataset, a large-scale benchmark for affective state recognition in e-learning. To enhance performance on underrepresented engagement classes, we apply targeted data augmentation techniques. Among the tested variants, ViBED-Net with LSTM achieves 73.43\% accuracy, outperforming existing state-of-the-art approaches. ViBED-Net demonstrates that combining face-aware and scene-aware spatiotemporal cues significantly improves engagement detection accuracy. Its modular design allows flexibility for application across education, user experience research, and content personalization. This work advances video-based affective computing by offering a scalable, high-performing solution for real-world engagement analysis. The source code for this project is available on \url{https://github.com/prateek-gothwal/ViBED-Net}.
\end{abstract}


\section{Introduction}
\label{sec1}
Detecting student engagement in online learning environments is a critical yet challenging task in the field of educational technology. As virtual classrooms become increasingly prevalent, the ability to monitor and respond to students’ attentiveness in real time has become essential for maintaining instructional quality and personalized learning. Engagement detection from video data, in particular, presents a rich but underexplored opportunity—not only in education, but across domains such as entertainment, marketing, and gaming, where understanding user attention and affective responses can provide actionable insights. For instance, content creators can leverage engagement detection to identify the most captivating segments of their films or games, enabling targeted improvements and audience personalization \cite{singh2023attention, b6}.

The emergence of datasets such as DAiSEE \cite{gupta2016daisee} and CMOSE \cite{wu2024cmose} has facilitated progress in this area by providing annotated video data for training and evaluating models. DAiSEE, in particular, has become a benchmark for developing affective state recognition systems in educational settings. Among the recent advancements, the method proposed by Malekshahi et al. \cite{malekshahi2024engagement} currently achieves state-of-the-art performance on the DAiSEE dataset by leveraging a generalizable deep learning framework capable of capturing both spatial and temporal cues from video data. This work surpasses earlier hybrid architectures such as EfficientNetB7 combined with TCN, LSTM, and Bi-LSTM \cite{selim2022efficientnet}, demonstrating enhanced generalization and accuracy in learner engagement detection.

In this paper, we present a novel two-stream architecture called ViBED-Net (Video Based Engagement Detection Network) that combines EfficientNetV2 with temporal modeling modules—Long Short-Term Memory (LSTM) networks and Transformer encoders—to detect student engagement from video with state-of-the-art performance on the DAiSEE dataset. Our approach processes both the facial region and the full video frame simultaneously, capturing fine-grained emotional cues and contextual scene information. The EfficientNetV2 backbone ensures efficient and high-resolution spatial feature extraction, while the LSTM and Transformer modules model temporal dependencies crucial for understanding engagement over time. This dual-stream and multi-temporal modeling strategy significantly outperforms existing benchmarks, demonstrating the effectiveness and flexibility of ViBED-Net for real-world engagement analysis applications.

By addressing a crucial gap in video-based affective computing, our work contributes a powerful, efficient, and extensible framework for multi-domain engagement analysis, with promising applications in online learning analytics, user experience research, and human-computer interaction.

\section{Literature Review}\label{sec2}
Student engagement detection has gained substantial traction in educational technology, especially with the widespread adoption of online learning platforms. A significant contribution to this domain is the DAiSEE dataset, which provides labeled videos capturing affective states such as engagement, boredom, confusion, and excitement in e-learning settings. As one of the earliest datasets of its kind, DAiSEE has served as a benchmark for numerous deep learning approaches targeting video-based engagement recognition \cite{gupta2016daisee}.

Building on this foundation, Selim et al. introduced a hybrid model combining EfficientNetB7 with Temporal Convolutional Networks (TCN), LSTM, and Bi-LSTM for engagement detection. Their approach, evaluated on the DAiSEE dataset, demonstrated strong performance by leveraging both spatial and temporal modeling capabilities \cite{selim2022efficientnet}. More recently, Malekshahi et al. proposed a general deep learning model that achieved state-of-the-art accuracy on DAiSEE by capturing engagement-related spatiotemporal patterns in a scalable and modular manner \cite{malekshahi2024engagement}.

Another influential work, \textit{Do I Have Your Attention?}, introduced a large-scale engagement dataset and established baseline models. The authors further validated their approach on DAiSEE, providing insights into cross-dataset generalization and the robustness of attention-based engagement predictors \cite{singh2023attention}.

In parallel, the CMOSE dataset emerged as a high-quality, multimodal resource that addresses limitations in labeling granularity and modality diversity. By incorporating eye gaze, posture, and facial expression modalities with synchronized high-quality labels, CMOSE enables more holistic modeling of student engagement \cite{wu2024cmose}.

Collectively, these studies underscore the importance of both spatial and temporal modeling in engagement prediction. However, many prior architectures focus primarily on either facial expressions or coarse global features, often neglecting the combined value of both streams. Our proposed model, ViBED-Net, addresses this gap through a dual-stream architecture that integrates EfficientNetV2 with both LSTM and Transformer-based temporal encoders. This allows it to simultaneously capture facial micro-expressions and full-scene context, achieving superior performance on DAiSEE and offering broader applicability across domains such as content creation, marketing analytics, and user experience research.

\section{Datasets}
\subsection{DAiSEE}
The DAiSEE (Dataset for Affective States in E-Environments) dataset is a large-scale video corpus curated to support research on affective computing in e-learning environments. It comprises 9,068 video snippets, each approximately 10 seconds long, collected from 112 participants in naturalistic online learning conditions. Each clip is annotated for four affective states—engaged, bored, confused, and excited—at four ordinal intensity levels: very low, low, high, and very high. Annotations were obtained via crowdsourcing and refined using majority voting to ensure label quality. The dataset includes diverse variations in lighting, facial occlusion, and head pose, making it suitable for training models that generalize to real-world conditions. Due to its scale, label richness, and ecological validity, DAiSEE has become a benchmark for video-based engagement detection in academic settings \cite{gupta2016daisee}.
\begin{figure}[ht]
\centering
\includegraphics[width=\linewidth]{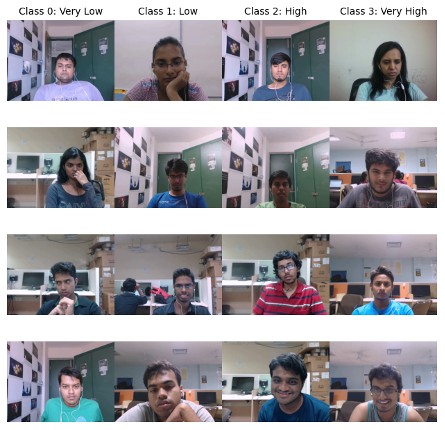}
\caption{Frames from DAiSEE dataset}
\label{fig:daisee_frames}
\end{figure}

\subsection{DAiSEE}
The DAiSEE (Dataset for Affective States in E-Environments) dataset is a large-scale video corpus curated to support research on affective computing in e-learning environments. It comprises 9,068 video snippets, each approximately 10 seconds long, collected from 112 participants in naturalistic online learning conditions. Each clip is annotated for four affective states—engaged, bored, confused, and excited—at four ordinal intensity levels: very low, low, high, and very high. Annotations were obtained via crowdsourcing and refined using majority voting to ensure label quality. The dataset includes diverse variations in lighting, facial occlusion, and head pose, making it suitable for training models that generalize to real-world conditions. Due to its scale, label richness, and ecological validity, DAiSEE has become a benchmark for video-based engagement detection in academic settings \cite{gupta2016daisee}.


\section{Proposed Methodology}
In this study, we introduce a dual-stream deep learning architecture designed to detect student engagement levels from video data. Our approach addresses two primary challenges: (1) the class imbalance in datasets like DAiSEE, particularly for “very low” and “low” engagement classes, and (2) the need to effectively model temporal dynamics inherent in video sequences.

To mitigate class imbalance, we employ targeted data augmentation techniques. These include salt-and-pepper noise, random translations, brightness and contrast adjustments, Gaussian blur, horizontal flips, and elastic transformations. Such augmentations have been shown to enhance model robustness and generalization, especially in scenarios with limited data for certain classes.

For feature extraction, we utilize EfficientNetV2, a convolutional neural network known for its balance between accuracy and computational efficiency. We extract two sets of features from each video frame: one from the entire frame to capture contextual information, and another from the cropped facial region to focus on fine-grained emotional cues. These features are precomputed and stored to expedite the training process.

The extracted features are then input into two separate temporal modeling modules. By default, we use Long Short-Term Memory (LSTM) networks, which are adept at capturing temporal dependencies in sequential data and modeling the progression of engagement levels over time. Additionally, we also experimented with a Transformer-based architecture as an alternative temporal encoder. Transformers, with their self-attention mechanism, are capable of capturing long-range dependencies and offer a parallelizable alternative to recurrent architectures. The outputs from these temporal models are concatenated and passed through a Multi-Layer Perceptron (MLP) to produce the final engagement level prediction.

\begin{figure*}[ht]
\centering
\includegraphics[width=\textwidth]{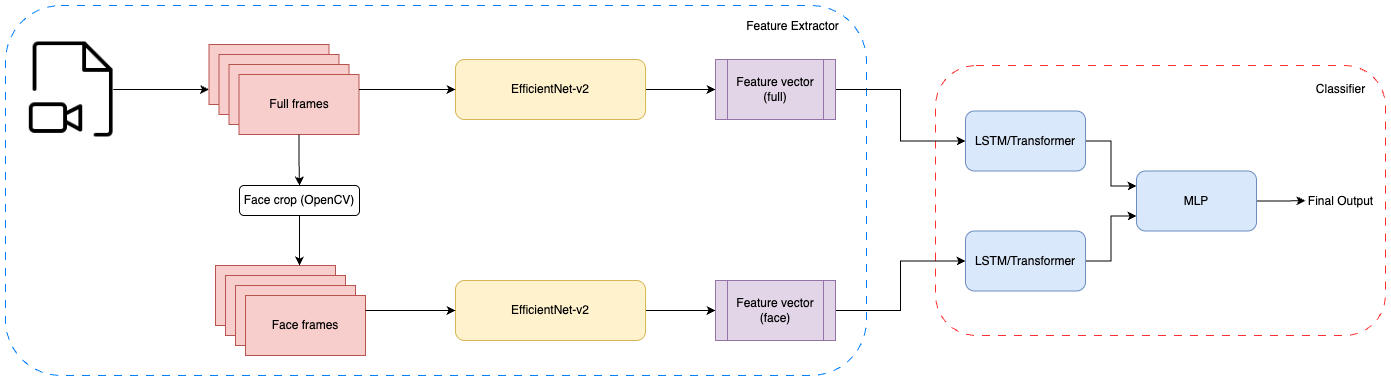}
\caption{ViBED-Net Architecture}
\label{fig:architecture}
\end{figure*}

\subsection{Model Architecture}
\subsubsection{Feature Extraction using EfficientNet}

EfficientNet is a family of convolutional neural networks introduced by Tan and Le \cite{b7}, known for achieving high accuracy with significantly fewer parameters and FLOPs compared to traditional architectures like ResNet or Inception. The core innovation behind EfficientNet is the compound scaling method, which uniformly scales the network’s depth, width, and input resolution using a set of fixed scaling coefficients. This enables more efficient use of model capacity and computational resources without requiring manual tuning of each dimension.

In our proposed architecture, we utilize EfficientNetV2, an enhanced version of the original that incorporates fused MBConv layers and progressive learning strategies for improved training speed and parameter efficiency \cite{b8}. Its balance between inference time and representational power makes it particularly suitable for large-scale video processing.

For each 10-second video clip in the DAiSEE dataset, we uniformly sample 60 frames to capture temporal variations in engagement. These frames are processed through two parallel EfficientNetV2 streams. The first stream receives the entire video frame as input, capturing global context such as posture, body language, and background. The second stream uses OpenCV-based face detection to crop the facial region from each frame, allowing the model to focus on fine-grained facial expressions and micro-movements crucial for affective analysis.

In both streams, we remove the classification head of EfficientNetV2 and extract features from the penultimate layer, resulting in a feature vector of size 1028 per frame. With 60 frames per video, this produces a final feature sequence of shape (60, 1028) for each stream. These feature arrays are precomputed and saved as pickle files to reduce computational overhead during training, allowing the LSTM components to operate directly on the extracted spatiotemporal representations.

Each of these feature sequences is then passed to a dedicated LSTM/Transformer network for temporal modeling, as described in the following subsection.

\subsubsection{Temporal Modeling using LSTM and Transformer}

While EfficientNetV2 extracts powerful spatial features from individual frames, engagement is inherently a temporal concept that unfolds over a sequence of video frames. To model this temporal evolution, we explore two distinct approaches: Long Short-Term Memory (LSTM) networks and a Transformer-based encoder. Both are designed to capture temporal dependencies, but with different mechanisms and computational trade-offs. In practice, we compare both variants during training to assess performance trade-offs in terms of accuracy, generalization, and computational cost.

\subsubsection{LSTM Architecture}

LSTMs are a type of recurrent neural network (RNN) well-suited for capturing long-term dependencies in sequential data. They mitigate the vanishing gradient problem of traditional RNNs through the use of internal memory cells and gating mechanisms—namely the input, forget, and output gates—which allow selective retention and updating of information over time \cite{b9}. This makes them highly effective for modeling the subtle progression of engagement cues across consecutive frames.

In our architecture, we use two independent LSTM networks: one for the full-frame features and one for the cropped facial region. Each LSTM receives a sequence of 60 feature vectors (one per frame), with an input size of 1028, corresponding to the output dimensionality from EfficientNetV2. The hidden size is set to 1024, with 1 layer and a dropout rate of 0.1 applied during training to prevent overfitting. The final hidden states from both LSTMs (one per stream) are extracted after processing the last time step and concatenated to form a unified representation for classification.

\subsubsection{Transformer Architecture}

In addition to LSTMs, we also experiment with a Transformer-based encoder, inspired by its success in various video understanding tasks \cite{vaswani2017attention}. Unlike recurrent networks, Transformers use a self-attention mechanism to weigh the importance of each frame in the sequence relative to all others. This allows the model to capture long-range temporal dependencies in parallel, making it both flexible and scalable.

Our Transformer model consists of:
\begin{itemize}
    \item A positional encoding module, which injects temporal order information into the input sequence (since Transformers lack inherent sequential bias) \cite{vaswani2017attention}.
    \item A stack of 2 Transformer encoder layers, each with 8 attention heads, a hidden dimension of 1024, and feed-forward layers of size 2048.
    \item Layer normalization and residual connections applied after each sub-layer to stabilize training \cite{vaswani2017attention}.
\end{itemize}

The input to the Transformer is the same $60 \times 1028$ sequence of features per stream. We apply separate Transformer encoders for the full-frame and face streams. After processing, the final token representations (either mean pooled or using a learnable [CLS]-like token) from each stream are concatenated and passed to the same MLP classifier as used with LSTMs.

\paragraph{Self-Attention Mechanism}
The core of the Transformer encoder is the scaled dot-product attention, which computes the relevance of each input frame (or token) to every other frame. Given a sequence of input feature vectors $X \in \mathbb{R}^{T \times d}$, where $T = 60$ is the number of frames and $d = 1028$ is the input dimension, the inputs are linearly projected into three matrices:

\begin{equation}
Q = XW^Q,\quad K = XW^K,\quad V = XW^V
\end{equation}

where $W^Q$, $W^K$, and $W^V$ are learnable projection matrices, and $Q$, $K$, and $V$ represent the queries, keys, and values respectively.

The self-attention output is computed as:
\begin{equation}
\text{Attention}(Q, K, V) = \text{softmax}\left( \frac{QK^\top}{\sqrt{d_k}} \right) V
\label{eq:self_attention}
\end{equation}

Here, $d_k$ is the dimensionality of the key vectors (typically $d_k = d / h$ for $h$ attention heads), and the softmax function is applied row-wise to generate attention weights that sum to 1.

\paragraph{Multi-Head Attention}
To capture diverse patterns from different subspaces, we use multi-head attention, which runs $h$ attention operations in parallel and concatenates their results:

\begin{equation}
\text{MultiHead}(Q, K, V) = \text{Concat}(\text{head}_1, \ldots, \text{head}_h) W^O
\end{equation}

\begin{equation}
\text{where} \quad \text{head}_i = \text{Attention}(QW_i^Q, KW_i^K, VW_i^V)
\end{equation}

$W^O$ is a learnable output projection matrix that combines the individual attention head outputs. This enables the model to learn relationships at multiple levels of granularity.

\paragraph{Feed-Forward Network.}
Each Transformer encoder block also includes a position-wise feed-forward network applied independently to each token:

\begin{equation}
\text{FFN}(x) = \max(0, xW_1 + b_1)W_2 + b_2
\end{equation}

Residual connections and layer normalization are applied after both the attention and feed-forward layers to ensure stable training.

\subsubsection{Classification using MLP}

After temporal modeling via the dual-stream LSTM networks, we obtain two fixed-length feature vectors—each of dimension 1024—representing the temporal summaries of the full-frame and face-based input sequences. These two vectors are concatenated to form a single combined feature vector of size 2048, capturing both scene-level and facial temporal dynamics in a unified representation.

\begin{figure}[ht]
\centering
\includegraphics[width=\linewidth]{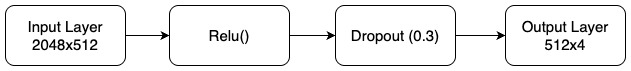}
\caption{MLP Architecture}
\label{fig:mlp}
\end{figure}

This concatenated vector is then passed to a Multi-Layer Perceptron (MLP) classifier to predict the final engagement level. The MLP consists of a fully connected layer that reduces the feature dimensionality from 2048 to 512, followed by a ReLU activation function and a dropout layer with a dropout rate of 0.3 to prevent overfitting. A final fully connected layer maps the 512-dimensional representation to the number of engagement classes (e.g., 4 in the DAiSEE dataset: Very Low, Low, High, Very High). The full MLP pipeline can be seen in Fig.~\ref{fig:mlp}.

This architecture allows the network to effectively combine and process both spatial and temporal features to classify student engagement levels with high accuracy. The entire classifier is trained using the cross-entropy loss function in conjunction with the AdamW optimizer, as described in the training section.

\subsection{Training Details}

The proposed architecture is trained in a supervised fashion using the cross-entropy loss, which is commonly applied to multi-class classification tasks. The loss for a single sample is calculated as shown in Equation~\eqref{eq:cross_entropy}:

\begin{equation}
\mathcal{L}{\text{CE}} = -\sum{i=1}^{C} y_i \log(\hat{y}_i)
\label{eq:cross_entropy}
\end{equation}

where $C$ is the number of classes, $y_i$ is the ground-truth one-hot encoded label, and $\hat{y}_i$ is the predicted probability for class $i$. This loss penalizes the model proportionally to the confidence in incorrect predictions.

For optimization, we use the AdamW optimizer, which improves upon Adam by decoupling the weight decay from the gradient update. The parameter update rule for each step $t$ is given in Equation~\eqref{eq:adamw}:

\begin{equation}
\theta_{t+1} = \theta_t - \eta \cdot \left( \frac{\hat{m}_t}{\sqrt{\hat{v}_t} + \epsilon} + \lambda \cdot \theta_t \right)
\label{eq:adamw}
\end{equation}

where $\eta$ is the learning rate (set to 0.001), $\hat{m}_t$ and $\hat{v}_t$ are the bias-corrected first and second moment estimates of the gradients, $\epsilon$ is a small constant for numerical stability, and $\lambda$ is the weight decay coefficient.

Training is performed using a mini-batch size of 32 over 40 epochs. To improve training efficiency, we precompute and cache the spatial features extracted from EfficientNetV2 as tensors of shape $60 \times 1028$ per video. These sequences are fed into the temporal and classification modules of the model.

The total loss minimized during training is the average cross-entropy loss over the $N$ training samples, as expressed in Equation~\eqref{eq:total_loss}:

\begin{equation}
\mathcal{L}{\text{total}} = \frac{1}{N} \sum{j=1}^{N} \mathcal{L}_{\text{CE}}^{(j)}
\label{eq:total_loss}
\end{equation}

Validation accuracy is monitored across epochs to ensure that the model generalizes well to unseen data.

\section{Results}
To evaluate the performance of our proposed engagement detection framework, we report the results using four standard classification metrics: Accuracy, Precision, Recall, and F1 score. These metrics provide a comprehensive understanding of how well the model performs, particularly in imbalanced class settings such as the DAiSEE dataset.

\subsection{Evaluation Metrics}

\begin{itemize}
    \item \textbf{Accuracy:} The proportion of total predictions that were correct.
    \begin{equation}
        \text{Accuracy} = \frac{TP + TN}{TP + TN + FP + FN}
    \end{equation}

    \item \textbf{Precision:} The proportion of positive identifications that were actually correct.
    \begin{equation}
        \text{Precision} = \frac{TP}{TP + FP}
    \end{equation}

    \item \textbf{Recall:} The proportion of actual positives that were correctly identified.
    \begin{equation}
        \text{Recall} = \frac{TP}{TP + FN}
    \end{equation}

    \item \textbf{F1-Score:} The harmonic mean of precision and recall, providing a balanced measure between the two.
    \begin{equation}
        \text{F1-Score} = 2 \cdot \frac{\text{Precision} \cdot \text{Recall}}{\text{Precision} + \text{Recall}}
    \end{equation}
\end{itemize}

These metrics are computed per class and averaged using a macro-average strategy to give equal weight to each class. This approach is especially useful for datasets with class imbalance.

\subsection{Quantitative Results}

\begin{table*}[ht]
\caption{Per-Class Performance Comparison of LSTM and Transformer Models}\label{tab:per_class_metrics}
\centering
\begin{tabular}{@{}lccccc@{}}
\toprule
\textbf{Model} & \textbf{Class} & \textbf{Prec} & \textbf{Rec} & \textbf{$F_1$} & \textbf{Acc} \\
\midrule
\multirow{4}{*}{LSTM} 
& Very Low & 1.00 & 0.25 & 0.40 & \multirow{4}{*}{0.73} \\
& Low      & 0.80 & 0.79 & 0.80 & \\
& High     & 0.77 & 0.68 & 0.72 & \\
& Very High& 0.70 & 0.79 & 0.74 & \\
\midrule
\multirow{4}{*}{Transformer} 
& Very Low & 0.00 & 0.00 & 0.00 & \multirow{4}{*}{0.62} \\
& Low      & 0.69 & 0.13 & 0.22 & \\
& High     & 0.60 & 0.76 & 0.67 & \\
& Very High& 0.66 & 0.53 & 0.59 & \\
\bottomrule
\end{tabular}
\end{table*}

Table~\ref{tab:per_class_metrics} presents a detailed per-class performance comparison between the LSTM-based and Transformer-based variants of ViBED-Net on the DAiSEE test set. The LSTM-based architecture achieves a superior overall accuracy of \textbf{73.43\%}, demonstrating its effectiveness in modeling the temporal dynamics of student engagement.

A closer look at the per-class metrics reveals important nuances. The LSTM model exhibits perfect precision (1.00) for the "Very Low" engagement class, meaning that every prediction it makes for this class is correct. However, its recall is only 0.25, indicating that it fails to identify 75\% of the actual "Very Low" instances. This suggests the model is conservative in its predictions for this minority class, prioritizing high confidence over high detection rates. In contrast, the model achieves a strong balance between precision and recall for the "Low" (0.80 F1-score) and "Very High" (0.74 F1-score) classes, indicating robust performance. The "High" engagement class shows slightly lower but still competitive performance with a 0.72 F1-score.

The Transformer-based model, with an overall accuracy of \textbf{62.39\%}, struggles significantly with the less-represented classes. It completely fails to identify any "Very Low" instances (0.00 precision and recall) and shows very low recall (0.13) for the "Low" engagement class. This suggests that the self-attention mechanism, without a recurrent structure, may be less effective at capturing the subtle and sparse temporal cues associated with disengagement in this dataset. While its performance on the "High" engagement class is reasonable (0.67 F1-score), its inability to handle class imbalance effectively makes it a less viable solution compared to the LSTM variant.

\begin{figure}[ht]
\begin{center}
\includegraphics[scale=0.5]{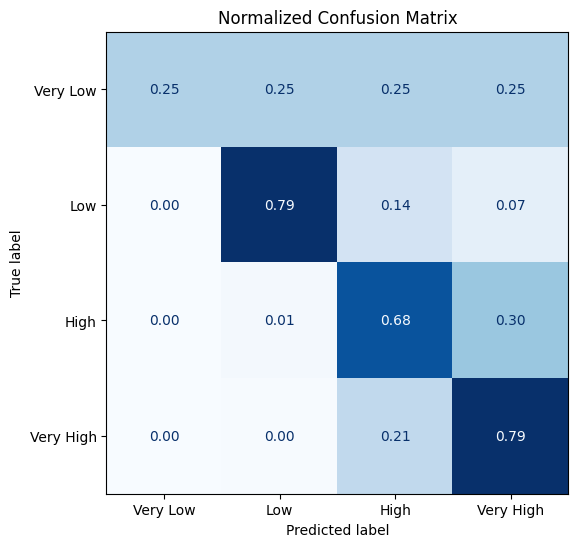}
\caption{Confusion matrix EfficientNet+LSTM on DAiSEE test set.}
\label{fig:confusion_matrix}
\end{center}
\end{figure}

These findings are further corroborated by the confusion matrix for the LSTM model, shown in Fig.~\ref{fig:confusion_matrix}, which visualizes the misclassifications between classes. The matrix clearly shows that most errors occur between adjacent engagement levels (e.g., "Low" vs. "High"), which is expected given the subjective nature of engagement annotation.

Finally, Table~\ref{tab:comparison_sota} situates our results within the broader landscape of existing methods. The ViBED-Net (LSTM) model not only surpasses the Transformer variant but also outperforms all previously reported state-of-the-art models on the DAiSEE dataset, establishing a new benchmark for video-based engagement detection.

\begin{table}[ht]
\caption{Accuracy Comparison with Existing Approaches on the DAiSEE Dataset}\label{tab:comparison_sota}
\centering
\begin{tabular}{@{}lc@{}}
\toprule
\textbf{Model} & \textbf{Accuracy (\%)} \\
\midrule
CNN-RNN Baseline (DAiSEE) \cite{gupta2016daisee} & 53.9 \\
I3D (Inflated 3D ConvNet) \cite{zhang2023cnn} & 52.35 \\
DFSTN (SE-ResNet-50 + LSTM + GALN) \cite{liao2021deep} & 58.84 \\
ResNet + TCN Hybrid \cite{abedi2021tcn} & 63.9 \\
EfficientNetB7 + LSTM \cite{selim2022efficientnet} & 67.48 \\
BiusFPN + ICCSA \cite{naveen2025biusfpn} & 68.16 \\
General Model for Learner Engagement \cite{malekshahi2024engagement} & 68.57 \\
Affect-driven Ordinal Engagement \cite{affectordinal2023} & 67.4 \\
VisioPhysioENet \cite{singh2024visiophysioenet} & 63.09 \\\midrule
Proposed ViBED-Net (Transformer) & 62.39 \\
\textbf{Proposed ViBED-Net (LSTM)} & \textbf{73.43} \\
\bottomrule
\end{tabular}
\end{table}

\section{Conclusion}\label{sec:conclusion}

In this paper, we introduced ViBED-Net, a novel dual-stream deep learning framework for video-based student engagement detection. By leveraging both face-aware and scene-aware spatiotemporal cues through parallel EfficientNetV2 feature extractors and LSTM-based temporal modeling, our approach achieved a new state-of-the-art accuracy of 73.43\% on the challenging DAiSEE dataset. Our results underscore the significant performance gains from combining fine-grained facial expression analysis with broader contextual information. The superior performance of the LSTM variant compared to the Transformer-based model suggests that recurrent architectures remain highly effective for modeling the nuanced temporal evolution of engagement in this context.

Despite its strong performance, our work has several limitations. First, while achieving high precision on the "Very Low" engagement class, the model's recall was low, indicating a difficulty in identifying all instances of this underrepresented class. This highlights the ongoing challenge of class imbalance in affective computing datasets. Second, our current framework relies on pre-computed features, which, while efficient for training, may not be ideal for fully real-time, end-to-end deployment. Finally, the model's robustness has been validated on a single dataset and assumes a relatively consistent, front-facing camera view, which may not hold true in all real-world learning environments.

Future research can extend this work in several promising directions. One key avenue is the integration of additional modalities to create a more holistic model of student engagement. Incorporating eye-gaze tracking, for instance, could provide a powerful, direct measure of visual attention. Furthermore, enhancing the model to accommodate different camera placements and viewing angles would significantly improve its practical applicability. This could be achieved by training on more diverse data or by incorporating view-invariant feature learning techniques. Finally, exploring self-supervised learning on large, unlabeled video datasets could help learn more generalizable representations of human behavior, which could then be fine-tuned for engagement detection, potentially improving cross-dataset performance and overall robustness.


\section*{ACKNOWLEDGMENT}
This material is based upon work supported by the National Science Foundation under Grant No. 2329919.

\bibliographystyle{plainnat}
\bibliography{references}

\end{document}